\documentclass{article}
\PassOptionsToPackage{numbers,sort&compress}{natbib}
\usepackage[preprint]{neurips_2023}
\usepackage[T1]{fontenc}
\usepackage[utf8]{inputenc}
\usepackage{amsmath,amssymb,booktabs,graphicx,array,tabularx,multirow}
\usepackage{xcolor,listings,microtype,url,needspace,placeins}
\usepackage{tikz}
\usetikzlibrary{arrows.meta,positioning,fit,calc}
\usepackage[colorlinks=true,linkcolor=blue!45!black,citecolor=blue!45!black,urlcolor=blue!45!black]{hyperref}
\definecolor{ink}{HTML}{153E54}
\definecolor{muted}{HTML}{536572}
\definecolor{pale}{HTML}{EDF3F6}
\makeatletter
\renewcommand{\@noticestring}{}
\makeatother
\newcommand{\code}[1]{\texttt{\detokenize{#1}}}

\newcommand{\reportauthors}{Xingyuan Zhang\\
Department of Chemistry, Wu Tsai Neurosciences Institute, Stanford University\\
\texttt{xingyuanzhang@stanford.edu}}
\newcommand{\projectlinks}{\href{https://github.com/straight-no-chaser/ProtLink}{Source repository}\quad\textperiodcentered\quad\href{https://huggingface.co/straight-no-chaser/ProtLink-Multispecies-HGT}{Released model}}
\title{ProtLink: Sequence and Multispecies Graph Context\\for Protein Interaction Prediction}
\author{\reportauthors}
\hypersetup{pdftitle={ProtLink: Sequence and Multispecies Graph Context for Protein Interaction Prediction},pdfauthor={Xingyuan Zhang}}
\begin{document}
\maketitle
\begin{abstract}
Predicting protein-protein interactions (PPIs) from sequence alone is difficult, while known interaction networks may provide informative biological context. We introduce ProtLink, a sequence-conditioned graph-learning framework that combines frozen ESM-2 representations with protein interaction topology and orthogroup-mediated connections between species. We compare sequence-only models, GraphSAGE, a homogeneous graph transformer, and a heterogeneous graph transformer (HGT) for human PPI prediction. In an initial edge-level comparison, three-layer multispecies HGT with ESM-2 650M features achieves an average precision (AP) of 0.980 under random negative sampling, compared with 0.951 for the homogeneous transformer; its AP decreases to 0.878 with two-hop-preferred negatives. In controlled three-seed experiments, three-layer full-graph HGT obtains an AP of $0.9701\pm0.0022$, exceeding an orthogroup-only model ($0.9435\pm0.0094$). The corresponding two-layer comparison shows little additional benefit from non-human PPI edges, consistent with the greater message-passing depth needed to transmit their information. For human proteins excluded from training interactions, multispecies HGT achieves $0.8335\pm0.0094$ AP versus $0.6482\pm0.0052$ for a sequence-only pair model. These results identify the value, depth dependence, and limits of evolutionary graph context while distinguishing prediction of held-out edges from generalization to held-out proteins.
\end{abstract}
\begin{center}\small\projectlinks\end{center}

\section{Introduction}
Protein-protein interactions (PPIs) organize signaling, molecular complexes, and the exchange of material in cells. A motivating example is biomolecular condensation, in which multivalent interactions recruit proteins into spatially organized assemblies. A longer-term goal is to extend interactome predictions to proteins with sparse network annotations, including engineered variants. Ranking plausible interaction partners could then help prioritize experimental follow-up. Here we study prediction of held-out database interactions as an initial benchmark toward that goal.

Protein sequences and graph neighborhoods offer complementary views. A sequence representation can be computed for a protein without known interactions. A graph encoder can summarize which interaction partners and biological contexts are already observed. Evolution adds another source of context: proteins in different species may belong to a shared orthogroup, allowing a model to exchange information without treating every interspecies pair as a physical interaction.

This study investigates these choices within a PyTorch Geometric (PyG) pipeline \citep{protlink,fey2019}. It compares a sequence-only pair MLP, GraphSAGE, a homogeneous graph transformer (GT), and multispecies HGT. The central question is when graph context adds predictive information beyond pretrained sequence features. A second question is how much the conclusion depends on negative sampling and the distinction between new edges and new proteins.

Our contribution is a sequence-conditioned, multispecies graph-learning framework and an evaluation of how architecture, embedding scale, and candidate selection affect human PPI prediction. Orthogroup nodes connect species-specific protein graphs, while supervised training remains restricted to human protein pairs. Controlled two- and three-layer ablations separate orthogroup membership from non-human interaction topology; a held-out-protein benchmark tests the extent to which multispecies graph context can support predictions without target-protein PPI edges in the training graph.

\section{Related work and model rationale}
ESM-2 learns representations from protein sequences \citep{lin2023}. We use its frozen, mean-pooled residue features as input rather than fine-tuning the language model. This makes the sequence-only baseline essential: gains from graph encoders should be evaluated against a model that sees the same sequence information.

GraphSAGE aggregates node features over local neighborhoods \citep{hamilton2017}, providing a baseline for incorporating known interaction topology. For attention-based message passing, we use PyG's \code{TransformerConv}, an operator also used in UniMP \citep{shi2021}; our link-prediction model does not use UniMP's label-propagation objective. HGT uses node-type- and relation-specific attention \citep{hu2020}, allowing the encoder to distinguish protein interactions from protein--orthogroup membership. Our implementation uses static heterogeneous message passing in PyG.

STRING integrates protein association evidence and provides a physical subnetwork \citep{szklarczyk2023}. Its physical links can include cocomplex evidence, so a positive label need not establish direct binary contact. Orthogroup inference provides an evolutionary framework for linking homologous protein families \citep{emms2019}, although orthology does not itself imply physical interaction. Evaluation also matters independently of architecture: pairwise prediction can share entities across training and test even when pairs differ \citep{park2012}. Independent experimental interaction maps, such as HuRI \citep{luck2020}, offer a route toward future validation outside the source network.

\section{Task and data}
\subsection{Graph and prediction target}
Let $V_P$ be proteins, $V_O$ orthogroups, and $H\subseteq V_P$ human proteins. The multispecies graph has node set $V=V_P\cup V_O$ and relation types
\begin{equation}
\mathcal{R}=\{\mathrm{interacts},\ \mathrm{in\_orthogroup},\ \mathrm{rev\_in\_orthogroup}\}.
\end{equation}
Interaction edges join proteins within the same species. Membership edges connect proteins to orthogroups and have explicit reverse edges. All proteins share the node type \code{protein}; a species identifier filters supervision but is not a separate node type in this implementation. The second node type is \code{orthogroup}.

Given the visible graph $G_{\mathrm{train}}$ and protein features $X$, the model assigns $p_\theta(u,v\mid G_{\mathrm{train}},X)$ to human pairs $u,v\in H$. For the edge-split benchmark, human training positives form the target-species message-passing edges. Human validation and test positives are excluded from that graph. Retained non-human PPI edges and orthogroup memberships remain available as context. The target task is therefore human PPI prediction conditioned on multispecies data, rather than prediction of physical interactions across species.

\subsection{Sources and preprocessing}
The single-species loader expects STRING v12.0 human physical links, a FASTA file, and cached ESM-2 features. It keeps links with \code{combined_score >= 700}, removes self-pairs, and deduplicates undirected pairs. It retains proteins with both sequence IDs and embedding IDs. The multispecies loader additionally intersects these IDs with protein metadata, verifies that each PPI's endpoints match the row's species, and retains proteins with an interaction or orthogroup context. The confidence score filters edges; it is not used as a message weight.

For each protein sequence $s_v$ with $L_v$ retained residues, features are
\begin{equation}
x_v=\frac{1}{L_v}\sum_{i=1}^{L_v}\operatorname{ESM2}(s_v)_i.
\end{equation}
Special tokens and padding are excluded. The embedding script can truncate at the tokenizer/model length limit, which must be recorded when regenerating features. The 35M and 650M variants correspond to different fixed feature spaces, not different GNN parameter counts. The released 650M configuration expects 1,280-dimensional protein features.

\section{Methods}
\subsection{Sequence and homogeneous graph baselines}
The pair MLP receives the ordered feature vector
\begin{equation}\label{eq:pair}
\phi(a,b)=[a\,\Vert\,b\,\Vert\,|a-b|\,\Vert\,a\odot b].
\end{equation}
It applies two hidden linear layers with ReLU and dropout, followed by a scalar logit. This is the sequence-only baseline when $a=x_u,b=x_v$. For graph models, $a=z_u,b=z_v$ are the encoder outputs. A dot decoder $z_u^\top z_v$ supplies an alternative with fewer parameters. The pair MLP is not invariant to exchanging its two endpoints, although the underlying PPI graph is undirected. We canonicalize the ordering of input pairs; a symmetric feature map or orientation-averaged prediction provides a useful decoder control.

The two-layer GraphSAGE encoder aggregates mean neighbor features. A layer can be written
\begin{equation}
h_v^{(\ell+1)}=W_{\mathrm{self}}^{(\ell)}h_v^{(\ell)}+
W_{\mathrm{nbr}}^{(\ell)}\operatorname{mean}_{u\in\mathcal{N}(v)}h_u^{(\ell)}+b^{(\ell)}.
\end{equation}
The implementation uses ReLU and dropout after the first layer. GT instead weights neighboring messages with multihead attention through \code{TransformerConv}, followed by LayerNorm, ReLU, dropout, and a residual addition when dimensions agree. It attends along visible graph edges, rather than across all protein pairs. Optional concatenation/addition of input features is also implemented.

\subsection{Multispecies HGT}
Protein inputs are projected into a hidden space, and orthogroups use learned embeddings:
\begin{equation}
h_v^{(0)}=W_Px_v+b_P\quad(v\in V_P),\qquad
h_o^{(0)}=e_o\quad(o\in V_O).
\end{equation}
HGT computes type-dependent queries, keys and values. For head $k$ and typed edge $u\xrightarrow{r}v$, a schematic attention rule is
\begin{align}
e_{uv}^{k}&=\frac{(q_v^k)^\top A_r^k k_u^k}{\sqrt{d_k}}\,\mu_{r,k},&
\alpha_{uv}^{k}&=\operatorname{softmax}_{u\in\mathcal{N}(v)}e_{uv}^{k},\\
m_v&=\mathop{\Vert}_{k}\sum_{u\xrightarrow{r}v}\alpha_{uv}^{k}M_r^k v_u^k.
\end{align}
Node-type-specific projections and HGT's output transformation produce the next representations. Each \code{HGTConv} is followed by LayerNorm, ReLU, dropout, and an outer residual connection. The main HGT architecture used in the initial model comparison has hidden width 256, \emph{three} layers, and four heads; we additionally evaluate two-layer HGT in the controlled ablation. A protein-orthogroup-protein path shares sequence-level information between species in two layers. By contrast, information originating from a non-human PPI neighbor must traverse a non-human interaction edge and two orthogroup-membership edges to reach a human protein, requiring at least three message-passing steps. This path-length difference motivates testing whether non-human PPI edges contribute differently at two and three layers.

\subsection{Learning and implementation}
For positive/negative training pairs $\mathcal{B}$, the objective is binary cross-entropy on logits $s_{uv}$:
\begin{equation}
\mathcal{L}=-\frac{1}{|\mathcal{B}|}\sum_{(u,v)\in\mathcal{B}}
\left[y_{uv}\log\sigma(s_{uv})+(1-y_{uv})\log(1-\sigma(s_{uv}))\right].
\end{equation}
Training resamples one negative per positive each epoch. Validation and test negatives are fixed for each seed, and test negatives exclude validation negatives. Candidate non-edges are screened against the observed positive network. Because negative labels arise from absent database edges rather than experimentally verified non-interactions, false-negative annotations remain possible; negative-list disjointness across training and evaluation is a separate sampling safeguard.

The graph encoders use full-graph forward passes, whereas the decoder evaluates batches of candidate pairs. Consequently, batching pair scores does not eliminate the encoder's memory requirements. Training uses Adam, learning rate $10^{-3}$, weight decay $10^{-5}$, dropout 0.2, a maximum of 30 epochs, early-stopping patience of 5, and seeds 0, 1, and 2. The best checkpoint maximizes validation AP. Appendix~\ref{app:code} provides representative PyG implementation excerpts.

\section{Evaluation protocol and evidence}
\subsection{Metrics and candidate distributions}
Average precision is the primary ranking metric,
\begin{equation}
\mathrm{AP}=\sum_j(R_j-R_{j-1})P_j,
\end{equation}
where $P_j,R_j$ are precision and recall along the ranked list. The F1 decision threshold, when used, is selected on the validation precision-recall curve and then held fixed for test. AP and ROC-AUC are threshold-independent. Evaluation uses equal numbers of observed positives and sampled non-edges; its precision and F1 do not directly represent the much lower prevalence among all possible human pairs. For the new controlled experiments we report AP for each of three seeds and mean $\pm$ sample standard deviation (SD), without inferring statistical significance from three repetitions.

Random negatives are uniformly sampled eligible non-edges. The hard sampler prefers non-edges connected by a two-hop path in the \emph{training} PPI graph and fills any shortfall with random non-edges. All known positives are excluded. However, the remaining unlabeled pairs are not necessarily experimentally verified non-interactions; the proportion of successful two-hop samples is needed to characterize the actual test distribution.

\subsection{Edge splits and node splits}
The default setting holds out approximately 10\% of positive edges for validation and 10\% for test. Its allocator first tries to preserve a training neighbor for every endpoint, then fills the requested counts if necessary. Thus ``edge random'' denotes a degree-aware randomized edge allocation, not unrestricted uniform partitioning.

We additionally examined a node-prioritized edge allocator. Because this procedure does not guarantee non-overlapping protein identities, it measures sensitivity to edge allocation rather than generalization to unseen proteins.

\subsection{Controlled orthogroup ablation}
To distinguish evolutionary context from additional interaction topology, we compare three graph configurations. \emph{Human-only} retains human training PPI edges; \emph{orthogroup-only} additionally contains non-human proteins and their orthogroup membership relations but no non-human PPI edges; \emph{full multispecies} also includes non-human PPI edges. Protein embeddings, human supervision, scoring task, and evaluation procedure follow the same ESM-2 650M benchmark design. Two-layer HGT is evaluated in all three configurations, and three-layer HGT in the orthogroup-only and full configurations. Each reported configuration has AP measurements for seeds 0, 1, and 2. We compare matched-depth models when assessing the contribution of non-human interactions, rather than treating an unpaired difference between network depths as a causal estimate.

\subsection{Inductive evaluation on held-out human proteins}
For the cold-protein benchmark, human proteins are partitioned by identity into seen and held-out subsets. Training positives connect seen proteins only; evaluation positives pair a held-out protein with a seen protein, and negative candidates are drawn from the same held-out-to-seen space. Held-out human PPI edges are excluded from message passing, while their sequence features and available orthogroup assignments remain accessible. This setup evaluates one-endpoint-inductive link prediction conditioned on sequence and orthology, not prediction for an entirely uncharacterized protein with no external context. We compare a sequence-only pair MLP with full multispecies HGT using seed-matched AP values for seeds 0, 1, and 2. The available summary reports AP, rather than per-protein or orthogroup-stratified outcomes.

\section{Results and analysis}
We first examine the model and candidate-sampling comparisons illustrated in Figures~\ref{fig:scale} and~\ref{fig:hard}, followed by the three-seed orthogroup and held-out-protein experiments in Figures~\ref{fig:ablation} and~\ref{fig:inductive}. The initial architecture comparison and the controlled experiments constitute distinct evaluation series; their AP values should be interpreted within the corresponding models, candidates, and training runs rather than pooled as a single replicate set.

\subsection{Graph context and decoder choice}
Under random negatives, the 35M sequence-only baseline has AP 0.723, compared with 0.922 for GraphSAGE with a dot decoder and 0.936 with the pair MLP decoder. The latter is a 0.213 AP difference relative to sequence-only inputs. The homogeneous transformer reaches 0.954, and multispecies HGT reaches 0.972. These comparisons are consistent with useful graph context. They do not prove that topology alone caused the gains, since feature filtering, decoder capacity, and graph coverage must also be held fixed.

The decoder comparison is not uniformly favorable to additional parameters. With 35M inputs, GraphSAGE + MLP exceeds GraphSAGE (dot) by 0.014 AP. With 650M inputs it is lower by 0.040 AP (0.893 versus 0.933). A flexible decoder may respond differently to the learned representation and training setup. Without matched run records and parameter-controlled comparisons, this is an interaction to investigate, rather than evidence that one decoder is universally superior.

\subsection{Embedding scale is architecture dependent}
\begin{figure}[t]\centering
\includegraphics[width=\linewidth]{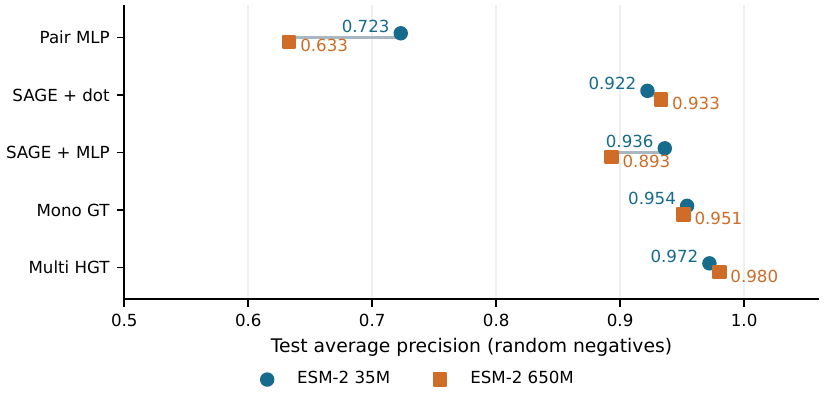}
\caption{Human PPI test average precision under the default edge split with randomly sampled non-edges. Points show individual model-comparison results for ESM-2 35M and 650M protein features; the HGT comparison uses a three-layer encoder. Lines connect embedding variants of the same model, rather than training epochs or replicate means. The axis begins at 0.5, the expected AP for random rankings under the balanced candidate distribution.}
\label{fig:scale}
\end{figure}

Figure~\ref{fig:scale} shows that increasing ESM-2 from 35M to 650M does not consistently improve downstream AP. Multispecies HGT gains 0.008 under random negatives and 0.023 under hard negatives. GraphSAGE (dot) gains 0.011 under random negatives. In contrast, pair MLP changes by $-0.090$, GraphSAGE + MLP by $-0.043$, and GT by $-0.003$ in the random setting. The records therefore support an architecture-dependent effect, not a monotonic advantage from larger embeddings. Possible explanations include optimization, different retained proteins, and overfitting.

\subsection{Hard negatives change the ranking}
\begin{figure}[t]\centering
\includegraphics[width=\linewidth]{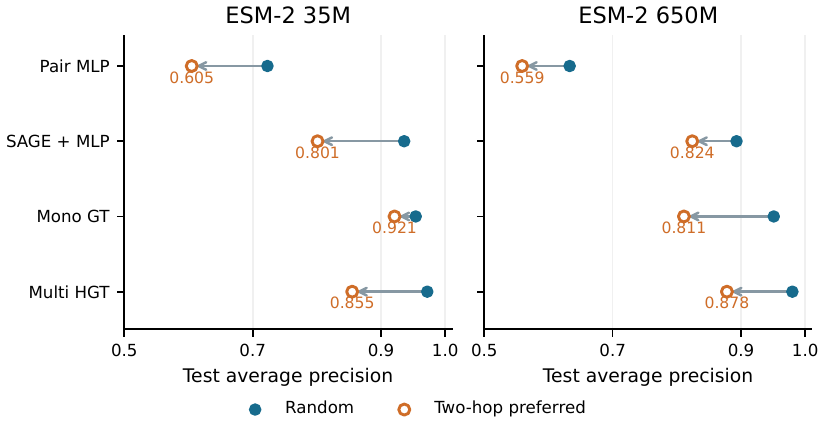}
\caption{Effect of negative-sampling difficulty on human PPI test AP. Each arrow connects the random-negative result (filled blue marker) with the two-hop-preferred negative result (open orange marker) for the same model and embedding size; HGT results use the three-layer encoder. The homogeneous graph transformer with ESM-2 35M yields the highest hard-negative AP in this comparison.}
\label{fig:hard}
\end{figure}

Moving from random to two-hop-preferred negatives reduces AP for every model with both results (Figure~\ref{fig:hard}). For 650M HGT, AP falls from 0.980 to 0.878, a 0.102 decrease. The corresponding declines are 0.074 for pair MLP, 0.069 for GraphSAGE + MLP, and 0.140 for GT. The smaller drop for a weaker model does not itself imply better robustness; absolute performance must be considered alongside sensitivity.

At 650M, multispecies HGT has the highest hard-negative AP among reported models: 0.878 versus 0.824 for GraphSAGE + MLP and 0.811 for GT. At 35M, however, GT reaches 0.921 and exceeds multispecies HGT's 0.855. This reversal is a substantive counterexample to an unconditional claim that multispecies HGT dominates. Two-hop candidate pairs share a local neighbor, making random non-edge discrimination an incomplete view of task difficulty.

\FloatBarrier

\subsection{Sensitivity to edge allocation}
Under the node-prioritized edge allocator, multispecies HGT reaches AP 0.956 and 0.968 with ESM-2 35M and 650M features, respectively. Relative to the default edge-split results, these values differ by $-0.016$ and $-0.012$. Because the two edge sets can share protein endpoints, this comparison measures sensitivity to edge allocation rather than generalization to unseen proteins.

\subsection{Contribution of orthogroup context}
\begin{figure}[t]\centering
\includegraphics[width=\linewidth]{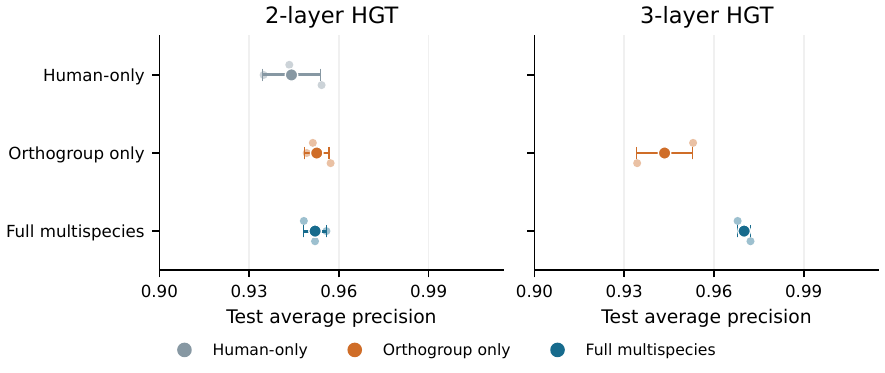}
\caption{Effect of multispecies graph relations on human PPI average precision at two HGT depths. \emph{Left}, two-layer HGT with human-only PPI context (gray), orthogroup membership and non-human protein features but no non-human PPI edges (orange), and the full multispecies graph (blue). \emph{Right}, three-layer HGT with orthogroup-only and full multispecies context; the human-only condition was not measured at this depth. Large markers and error bars show mean $\pm$ sample SD across seeds 0, 1, and 2; smaller translucent markers show individual seeds.}
\label{fig:ablation}
\end{figure}

The ablation separates two distinct mechanisms of multispecies context (Figure~\ref{fig:ablation}). With \emph{two} HGT layers, the human-only model achieves test AP $0.9442\pm0.0097$ (mean $\pm$ SD across three seeds). Adding orthogroup membership and non-human sequence features, but no non-human PPI edges, gives $0.9526\pm0.0042$ AP, an average paired improvement of $0.0084$. The full multispecies graph scores $0.9520\pm0.0038$, only $-0.0005$ AP relative to the orthogroup-only model on average. The seed-level full-minus-orthogroup differences range from $-0.0052$ to $+0.0066$, so the two-layer comparison provides no consistent benefit from non-human PPI edges.

Increasing HGT to \emph{three} layers changes this relationship. Orthogroup-only HGT obtains $0.9435\pm0.0094$ AP, whereas full multispecies HGT reaches $0.9701\pm0.0022$ AP. The paired mean improvement is $0.0266$ AP, positive in all three seeds ($0.0149$, $0.0270$, and $0.0379$). A three-step path can transmit information from a non-human interaction neighbor to a human target through orthogroup membership, whereas two layers can transmit orthogroup-mediated protein information but not that PPI-neighbor signal. The observations are consistent with this receptive-field explanation, although they do not isolate path length from all other effects of adding an HGT layer.

\subsection{Generalization to held-out human proteins}
\begin{figure}[t]\centering
\includegraphics[width=\linewidth]{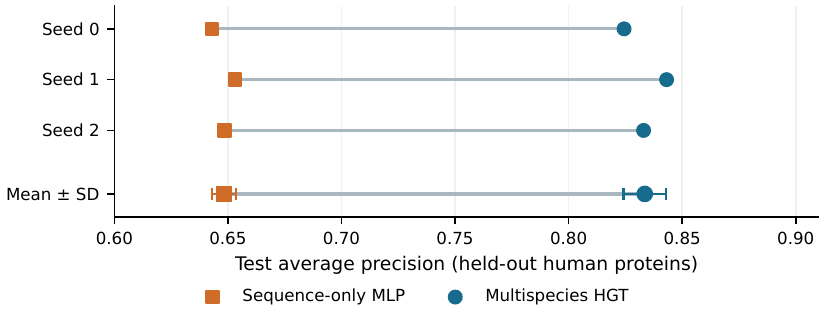}
\caption{One-endpoint-inductive human PPI prediction using held-out-to-seen candidate pairs. Orange squares represent the sequence-only pair MLP; blue circles represent multispecies HGT. Connected pairs show test AP for the same seed (0, 1, or 2). The bottom row reports mean $\pm$ sample SD across seeds. Protein sequence features and available orthogroup context are retained for held-out proteins, while their supervised human PPI edges are withheld from the training graph.}
\label{fig:inductive}
\end{figure}

The one-endpoint-inductive experiment evaluates candidate interactions in which the held-out human protein has no supervised PPI edges in the training graph. The sequence-only pair MLP reaches test AP $0.6482\pm0.0052$, while full multispecies HGT reaches $0.8335\pm0.0094$ (Figure~\ref{fig:inductive}). The mean paired difference is $0.1853$ AP, with gains of $0.1815$, $0.1899$, and $0.1844$ in seeds 0, 1, and 2. This gain is substantially larger than the within-architecture variation across these three seeds.

Held-out proteins may still be represented by sequence embeddings and orthogroup connections to proteins whose interactions are observed. Accordingly, this comparison supports generalization beyond supervised \emph{human PPI edges} for the held-out target proteins, but does not establish performance on de novo sequences without homologous network context. The aggregate results do not resolve how much of the gain is concentrated among proteins with informative orthogroup memberships, motivating a more stringent orthogroup-disjoint evaluation.

\FloatBarrier

\section{Discussion and limitations}
\paragraph{Biological interpretation.}
Orthogroup membership provides an evolutionary prior without implying physical contact between species. At two HGT layers, orthogroup-only context improves on human-only context, whereas adding non-human PPI edges has no consistent effect. At three layers, these edges improve AP by $0.0266$, consistent with the three-hop route from a non-human PPI neighbor to a human protein. The association with message-passing depth does not, by itself, establish the mechanism of biological transfer. Since orthology-derived evidence can also contribute to interaction databases, provenance-filtered links, temporal holdouts, and independently assayed interactions would provide stronger tests.

\paragraph{New proteins and engineered sequences.}
Removing the held-out proteins' human PPI edges still leaves their sequences and orthogroups visible to HGT. Its $0.1853$ mean AP gain therefore supports conditional inductive prediction, not independence from previously characterized homologs. Sequence-cluster and orthogroup-disjoint partitions would test stronger generalization. Engineered chimeras, mutants, and truncated proteins may not map to established orthogroups.

\paragraph{Computational and statistical limits.}
Full-graph message passing scales poorly: $N$ float32 embeddings of dimension $d$ occupy $4Nd$ bytes before edges, gradients, and activations. Neighborhood sampling or graph partitioning may be required at larger scale. Controlled results include only three seeds and aggregate AP, limiting statistical inference and per-protein stratification. Finally, the balanced positive/non-edge benchmark differs from the low-prevalence setting of screening all possible human PPI pairs.

\section{Conclusion}
ProtLink integrates protein sequence representations, graph message passing, and orthogroup-based multispecies context for human PPI prediction. In the initial edge-level comparison, three-layer HGT reaches AP 0.980 with random negatives and 0.878 with two-hop-preferred negatives, illustrating the dependence of model ranking on candidate difficulty. Controlled experiments show a $0.0266$ mean AP gain from adding non-human PPI topology to three-layer HGT, compared with essentially no change at two layers. For interactions involving held-out human proteins, full multispecies HGT achieves $0.8335\pm0.0094$ AP versus $0.6482\pm0.0052$ for the sequence-only baseline. Taken together, the findings support depth-dependent use of evolutionary network context while emphasizing that strong database link-prediction results are not a substitute for experimental validation on genuinely novel proteins.

\clearpage\bibliographystyle{plainnat}
\bibliography{references}

@inproceedings{hamilton2017,
 title={Inductive Representation Learning on Large Graphs},
 author={Hamilton, William L. and Ying, Rex and Leskovec, Jure},
 booktitle={Advances in Neural Information Processing Systems}, volume={30}, year={2017},
 url={https://arxiv.org/abs/1706.02216}}

@inproceedings{hu2020,
 title={Heterogeneous Graph Transformer},
 author={Hu, Ziniu and Dong, Yuxiao and Wang, Kuansan and Sun, Yizhou},
 booktitle={Proceedings of The Web Conference 2020},pages={2704--2710},year={2020},
 doi={10.1145/3366423.3380027},url={https://arxiv.org/abs/2003.01332}}

@article{lin2023,
 title={Evolutionary-scale prediction of atomic-level protein structure with a language model},
 author={Lin, Zeming and Akin, Halil and Rao, Roshan and Hie, Brian and Zhu, Zhongkai and Lu, Wenting and Smetanin, Nikita and Verkuil, Robert and Kabeli, Ori and Shmueli, Yaniv and Costa, Allan dos Santos and Fazel-Zarandi, Maryam and Sercu, Tom and Candido, Salvatore and Rives, Alexander},
 journal={Science},volume={379},number={6637},pages={1123--1130},year={2023},doi={10.1126/science.ade2574},
 url={https://www.science.org/doi/10.1126/science.ade2574}}

@article{szklarczyk2023,
 title={The {STRING} database in 2023: protein--protein association networks and functional enrichment analyses for any sequenced genome of interest},
 author={Szklarczyk, Damian and Kirsch, Rebecca and Koutrouli, Mikaela and Nastou, Katerina and Mehryary, Farrokh and Hachilif, Radja and Gable, Annika L. and Fang, Tao and Doncheva, Nadezhda T. and Pyysalo, Sampo and Bork, Peer and Jensen, Lars J. and von Mering, Christian},
 journal={Nucleic Acids Research},volume={51},number={D1},pages={D638--D646},year={2023},doi={10.1093/nar/gkac1000}}

@article{park2012,
 title={A flaw in the typical evaluation scheme for pair-input computational predictions},
 author={Park, Yungki and Marcotte, Edward M.},journal={Nature Methods},volume={9},number={12},pages={1134--1136},year={2012},doi={10.1038/nmeth.2259},
 url={https://pmc.ncbi.nlm.nih.gov/articles/PMC3531800/}}

@article{emms2019,
 title={{OrthoFinder}: phylogenetic orthology inference for comparative genomics},
 author={Emms, David M. and Kelly, Steven},
 journal={Genome Biology},volume={20},pages={238},year={2019},
 doi={10.1186/s13059-019-1832-y},
 url={https://genomebiology.biomedcentral.com/articles/10.1186/s13059-019-1832-y}}

@article{luck2020,
 title={A reference map of the human binary protein interactome},
 author={Luck, Katja and Kim, Dae-Kyum and Lambourne, Leigh and Spirohn, Katherina and Begg, Beatrix E. and Bian, Wei and Brignall, Robert and Cafarelli, Travis and Campos-Laborie, Francisco J. and Charloteaux, Benoit and others},
 journal={Nature},volume={580},pages={402--408},year={2020},
 doi={10.1038/s41586-020-2188-x},
 url={https://www.nature.com/articles/s41586-020-2188-x}}

@misc{fey2019,
 title={Fast Graph Representation Learning with {PyTorch Geometric}},
 author={Fey, Matthias and Lenssen, Jan Eric},year={2019},howpublished={ICLR Workshop on Representation Learning on Graphs and Manifolds},
 url={https://arxiv.org/abs/1903.02428}}

@inproceedings{shi2021,
 title={Masked Label Prediction: Unified Message Passing Model for Semi-Supervised Classification},
 author={Shi, Yunsheng and Huang, Zhengjie and Feng, Shikun and Zhong, Hui and Wang, Wenjin and Sun, Yu},
 booktitle={Proceedings of IJCAI},year={2021},pages={1548--1554},url={https://arxiv.org/abs/2009.03509}}

@misc{pygdocs,
 author={{PyG Team}},title={{PyTorch Geometric} documentation},year={2026},
 howpublished={SAGEConv, TransformerConv, HGTConv, and heterogeneous graph learning documentation},
 url={https://pytorch-geometric.readthedocs.io/en/latest/notes/heterogeneous.html},note={Accessed November 5, 2026}}

@misc{protlink,
 author={{ProtLink project}},title={{ProtLink}: Multispecies {HGT} for interactome prediction},year={2026},
 url={https://github.com/straight-no-chaser/ProtLink},note={Accompanying source snapshot and result figures}}

\clearpage\appendix
\section{PyG implementation walkthrough}\label{app:code}
These excerpts use the current repository API. The notebook provides a synthetic demonstration and an optional real-data path. PyG represents typed adjacency through \code{HeteroData} and consumes it with \code{HGTConv} \citep{pygdocs}.

\subsection{Construct the visible graph}
The repository's builder adds non-human interactions to training human interactions. Passing \code{train_pos} rather than all human positives is essential: otherwise the encoder would see the target edges it is asked to predict.
\begin{lstlisting}
from protlink.multispecies_graph_utils import build_multispecies_hetero_data
data, metadata = build_multispecies_hetero_data(dataset, train_pos)
assert data.node_types == ["protein", "orthogroup"]
print(data.metadata())  # Two node types and three directed relations.
\end{lstlisting}
Both membership directions are required so that protein and orthogroup nodes can receive messages. PPI pairs are also expanded into both directions for propagation. Candidate pairs for the decoder remain one copy of each undirected pair.

\subsection{Encode and score human candidates}
\begin{lstlisting}
from protlink.models import HGTEncoder, PairMLPDecoder

encoder = HGTEncoder(
    protein_input_dim=data["protein"].x.shape[1],
    hidden_dim=256, heads=4, layers=3, dropout=0.2,
    metadata=data.metadata(),
    node_type_counts={"orthogroup": data["orthogroup"].num_nodes},
)
decoder = PairMLPDecoder(256, hidden_dim=256, dropout=0.2)
z = encoder(data)
logits = decoder(z, human_pairs)  # LongTensor, shape [B, 2].
\end{lstlisting}
\code{HGTEncoder} returns protein representations. Decoder batching does not reduce the full-graph encoder's memory requirement.
The two-layer ablation changes \code{layers} to \code{2} without modifying the prediction target.

\subsection{Separate node identities before sampling negatives}
\begin{lstlisting}
from protlink.split import split_positive_edges
from protlink.negative_sampling import sample_negative_edges

split = split_positive_edges(
    dataset["edges"], dataset["num_nodes"], seed=0, mode="node_inductive",
    candidate_nodes=dataset["target_node_indices"],
)
assert not set(split["train_nodes"]) & set(split["test_nodes"])
test_neg = sample_negative_edges(
    dataset["num_nodes"], len(split["test_pos"]),
    dataset["edges"], seed=0,
    left_candidate_nodes=split["test_nodes"],
    right_candidate_nodes=split["train_nodes"])
\end{lstlisting}
Positives and negatives must share the same pair space. Held-out nodes retain sequence features and orthogroup context. Two-hop candidate selection is not equivalent to random negative sampling under this split, because held-out proteins have no training human PPI neighbors.

\clearpage\section{Experimental reproducibility}\label{app:provenance}
\subsection{Result files and checkpoints}
The numerical results underlying Figures 1-4 are provided in the accompanying \code{results.csv} file. The dataset contains test average precision values for the architecture comparisons and seed-level measurements, means, and standard deviations for the controlled ablation and held-out-protein experiments. Reproducing the computational experiments requires the corresponding protein sequences, interaction networks, orthogroup assignments, model configurations, data splits, and fixed evaluation candidates.

\subsection{Training and evaluation}
The training entry point is \code{python -m protlink.train graph}. With aligned sequence, metadata, graph, and embedding files, a representative multispecies command is:
\begin{lstlisting}[language=bash]
python -m protlink.train graph \
  --experiment multispecies_hetero --graph-type hetero \
  --encoder hgt --decoder mlp \
  --fasta data/proteins.fasta \
  --protein-metadata data/protein_metadata.tsv \
  --ppi-edges data/ppi_edges.tsv \
  --protein-to-orthogroup data/protein_to_orthogroup.tsv \
  --embeddings data/esm2_t33_650m.npz \
  --target-species 9606 --seeds 0 1 2 \
  --negative-mode random --split-mode edge_random \
  --device cuda --save-preds --output-dir outputs/reproduction
\end{lstlisting}
The negative-sampling setting can be changed to \code{two_hop_hard}; \code{node_inductive} selects the held-out-protein protocol. Replication requires the same STRING release, embedding and truncation settings, edge split, negative candidate lists, software versions, and random seeds as the reported evaluation.

\subsection{Split validation}
Five unit tests in \code{tests.test_node_inductive_split} check the allocation of held-out node identities, the endpoints of held-out-to-seen pairs, negative-sampling eligibility, and expected return types. These synthetic tests validate the split implementation; predictive performance is evaluated separately on the STRING-derived interaction benchmark.
\end{document}